# Generalist Large Language Models Outperform

# Clinical Tools on Medical Benchmarks

Krithik Vishwanath[1-3], Mrigayu Ghosh[4,5], Anton Alyakin, M.S.E.[1,6,7] ,

Daniel Alexander Alber, B.S.[1] , Yindalon Aphinyanaphongs, M.D., Ph.D.[7-9],

Eric Karl Oermann M.D.[1,7,10,11]

[1]Department of Neurological Surgery, NYU Langone Health, New York, New York, USA

[2]Department of Aerospace Engineering & Engineering Mechanics, The University of Texas at Austin, Austin, Texas, USA

[3]Department of Mathematics, The University of Texas at Austin, Austin, Texas, USA

[4]Department of Biomedical Engineering, The University of Texas at Austin, Austin, Texas, USA

[5]Department of Molecular Biosciences, The University of Texas at Austin, Austin, Texas, USA

[6]Department of Neurosurgery, Washington University School of Medicine in St. Louis, St. Louis, Missouri, USA

[7]Global AI Frontier Lab, New York University, New York, New York, USA

[8]Department of Population Health, NYU Langone Health, New York, New York, USA

[9]Department of Medicine, NYU Langone Health, New York, New York, USA

[10]Department of Radiology, NYU Langone Health, New York, New York, USA

[11]Center for Data Science, New York University, New York, New York, USA

**Correspondence:**

Krithik Vishwanath

Department of Neurosurgery,

NYU Langone Medical Center,

New York University, 550 First Ave, MS 3 205,

New York, NY10016, USA.

Email: krithik.vish@utexas.edu

Eric K. Oermann, MD

Department of Neurosurgery,

NYU Langone Medical Center,

New York University, 550 First Ave, MS 3 205,

New York, NY10016, USA.

Email: eric.oermann@nyulangone.org

**Abstract**

Specialized clinical AI assistants are rapidly entering medical practice, often framed as safer or more reliable than general-purpose large language models (LLMs). Yet, unlike frontier models, these clinical tools are rarely subjected to independent, quantitative evaluation, creating a critical evidence gap despite their growing influence on diagnosis, triage, and guideline interpretation. We assessed two widely deployed clinical AI systems (OpenEvidence and UpToDate Expert AI) against three state-of-the-art generalist LLMs (GPT-5, Gemini 3 Pro, and Claude Sonnet 4.5) using a 1,000-item mini-benchmark combining MedQA (medical knowledge) and HealthBench (clinician-alignment) tasks. Generalist models consistently outperformed clinical tools, with GPT-5 achieving the highest scores, while OpenEvidence and UpToDate demonstrated deficits in completeness, communication quality, context awareness, and systems-based safety reasoning. These findings reveal that tools marketed for clinical decision support may often lag behind frontier LLMs, underscoring the urgent need for transparent, independent evaluation before deployment in patient-facing workflows.

**Introduction**

Large language models (LLMs) are increasingly used in clinical medicine.[1–3] While major frontier models are well-benchmarked[1,2], specialized tools marketed for clinical use remain under-studied. OpenEvidence, claimed to be used by 40% of U.S. physicians[3] and to ace the USMLE[4], has only been evaluated either internally[4] or qualitatively[5]. Similarly, the newly released UpToDate Expert AI, already used by clinicians at an estimated 70% of major enterprise health systems[6], still awaits any sort of independent quantitative evaluation. There is a substantial gap between marketing claims and rigorous, peer-reviewed assessment of clinical AI performance. This has created a situation in which individual clinicians and health systems may adopt AI systems without adequate evidence regarding their limitations, biases, comparative performance, or failure modes. In settings where AI outputs influences diagnosis or management decisions, this lack of transparency may introduce avoidable clinical risk and challenges evidence-based integration of AI into care.

It is commonly hypothesized that specialist tools may be superior to generalist LLMs through domain-specific training[7–9] or retrieval augmented generation (RAG)[3,6,7]. However, specialist systems vary in their design, scope, and implementation. Frontier generalist models benefit from substantially larger training corpora and more advanced alignment, which may enable them to match or exceed specialist performance depending on the task. Independent evaluations capable of comparing these two categories of models on standardized, clinically relevant benchmarks are critical to clarify whether specialization offers measurable advantage. To address this gap, we conducted the first quantitative assessment of specialized clinical AI tools (OpenEvidence, UpToDate Expert AI) and leading generalist models (GPT-5, Gemini 3 Pro Preview, and Claude

Sonnet 4.5) on a mini-benchmark comprising knowledge-intensive (MedQA[1]) and expert-alignment (HealthBench[2]) tasks.

## Methods

We created a medical mini-benchmark by randomly sampling 500 USMLE-style questions from MedQA[1] to assess factual knowledge, and 500 prompts from HealthBench[2] to evaluate alignment with expert judgement. The resulting 1,000 questions were used to assess GPT-5 (snapshot gpt-5-2025-08-07), Gemini 3 Pro Preview (accessed 11/20/25), and Sonnet 4.5 (snapshot claude-sonnet-4-5-20250929) via their respective APIs, while OpenEvidence (accessed 09/25) and UpToDate Expert AI (accessed 11/25) were manually queried via a browser interface. OpenEvidence and UpToDate were grouped as clinical tools, whereas the other three (GPT-5, Gemini 3 Pro, and Sonnet 4.5) were considered to be generalist models.

HealthBench responses were graded for the proportion of rubric points achieved across five axes: accuracy, completeness, communication quality, context awareness, and instruction following. HealthBench questions were restricted to prompts requiring a single, standalone response (i.e., no multi-turn interactions). Responses were also subset into seven themes: emergency referrals, context seeking, global health, health data tasks, expertise-tailored communication, responding under uncertainty, and response depth.

Scoring for HealthBench and MedQA were completed by GPT-4.1 (gpt-4.1-2025-04-14) which demonstrated reliability in grading medical Q&A[2]. Holm–Bonferroni-adjusted statistical significance was established using exact McNemar's and Wilcoxon's tests. For grouped comparisons of clinical tools to generalist models, statistical significance was established using Welch's *t* tests and Šidák-adjusted two-way ANOVA.

**Results**

To evaluate model performance on medical question answering, we first assessed accuracy on MedQA. The resultant MedQA accuracies were: GPT-5, 96.2% (95% CI, 94.1%–97.6%); Gemini, 94.6% (92.3%–96.3%); Sonnet 4.5, 91.4% (88.6%–93.6%); OpenEvidence, 89.6% (86.6%–92.0%); and UpToDate, 88.4% (85.3%–90.9%); with GPT-5 outperforming all models except Gemini (McNemar $P<1\times10^{-4}$ vs. OpenEvidence and vs. UpToDate, $P=0.0008$ vs. Sonnet). Gemini outperformed OpenEvidence ($P=0.003$), UpToDate ($P<1\times10^{-4}$), and Sonnet ($P=0.026$). Other differences were not significant (GPT-5 vs. Gemini, $P=0.51$; OpenEvidence vs. Sonnet, $P=0.52$; UpToDate vs. Sonnet, $P=0.21$; OpenEvidence vs. UpToDate, $P=0.52$) (**Figure 1A**). OpenEvidence, UpToDate Expert AI, and Sonnet 4.5 perform worse on *Medical Knowledge / Scientific Topic* questions ($P<0.008$; **Supplemental Figure 1**).

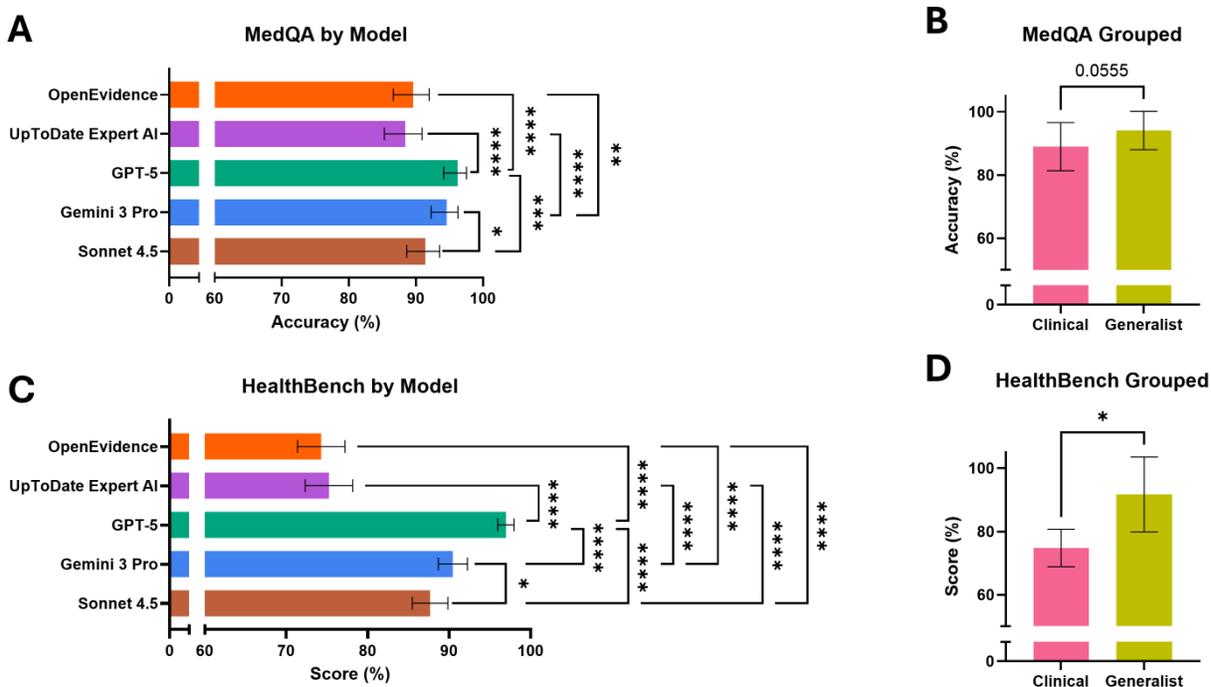

**Figure 1. Performance of generative artificial intelligence (AI) tools on medical benchmarks. A)** MedQA accuracy by model. Statistical significance established via exact McNemar's tests. **C)** HealthBench average scores by model, assessed through a physician-consensus rubric. Statistical significance established via pairwise Wilcoxon's tests. **B, D)** Comparison of the performance of Clinical (Expert AI, Open Evidence) tools to Generalist (GPT 5, Gemini 3 Pro, and Sonnet 4.5) models on MedQA and HealthBench, respectively. Statistical significance established via Welch's *t*-tests. Error bars represent 95% confidence intervals in A-D.  \* $P < 0.05$, \*\* $P < 0.01$, \*\*\* $P < 0.001$, \*\*\*\* $P < 0.0001$.

When comparing clinical tools to generalist models in a grouped fashion, generalist models (average accuracy of 94.1%) outperformed clinical tools (average accuracy of 89.0%), although this difference was not significant (*P*=0.056; **Figure 1B**). Notably, when questions are classified by their competency category, clinical tools perform worse on *Systems-based Practice Questions, including Patient Safety* ($P<1\times10^{-4}$; **Supplemental Figure 2**).

To assess model agreement with expert clinicians, we evaluate performance on HealthBench. HealthBench mean consensus scores were: GPT-5, 97.0% (96.0–98.0%); Gemini, 90.5% (88.7–92.3%); Sonnet, 87.7% (85.5–89.9%); UpToDate, 75.2 (72.3–78.1%); OpenEvidence, 74.3% (71.4–77.2%). GPT-5 outperformed other models (all Wilcoxon $P<1\times10^{-9}$), while OpenEvidence consistently underperformed ($P<1\times10^{-10}$), except when compared to UpToDate ($P=0.58$) (**Figure 1C**). Furthermore, the generalist models outperformed the clinical tools by about 1.23-fold (91.7% vs. 74.8%; $P=0.023$; **Figure 1D**).

In axis-level analysis, GPT-5 achieved the highest or tied-for-highest score across all five axes (**Figure 2A**). OpenEvidence consistently ranked lowest or tied for lowest with UpToDate, with all GPT-5 vs. OpenEvidence and GPT-5 vs. UpToDate comparisons being significant (McNemar $P\le5.5\times10^{-4}$). In three out of five axes, OpenEvidence was inferior to all three generalist models ($P\le5.8\times10^{-4}$), and in two out of five axes, UpToDate was inferior to all three generalist models ($P\le2.7\times10^{-4}$). These trends correspond with the grouped clinical vs. generalist model analysis, which demonstrated superiority of generalist models in completeness ($P<10^{-4}$), communication quality ($P=0.002$), and context awareness ($P=0.012$) (**Figure 2C**).

In a theme-level analysis, GPT-5 ranked first or tied for first in all categories, with perfect scores in four. OpenEvidence and UpToDate scored lowest or tied for lowest in all seven themes (**Figure 2B**). Again, all GPT-5 vs. OpenEvidence and GPT-5 vs. UpToDate comparisons were significant (Wilcoxon *P*≤0.005). Grouped analysis revealed that there were significant differences in the scores of clinical tools and generalist models when responding to questions under the cluster themes of emergency referrals (*P*=0.0002) and expertise-tailored communication (*P*=0.0392) (**Figure 2D**).

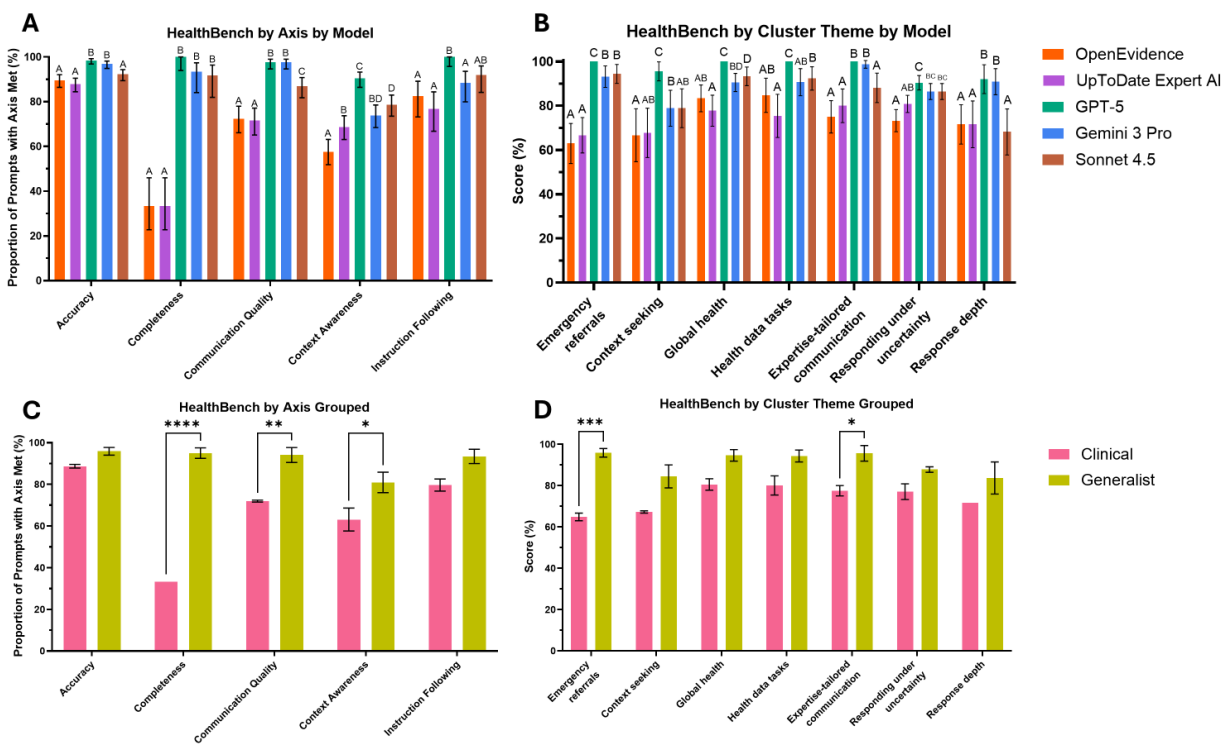

**Figure 2. Performance of generative AI tools on HealthBench by axis and by theme. A)** Proportion of prompts with axis met in HealthBench by model. Statistical significance established via pairwise exact McNemar's tests. **B)** Mean score in HealthBench by cluster theme by model. Statistical significance established via pairwise Wilcoxon's tests. In A-B, error bars represent 95% confidence intervals, and labeled letters indicate significant differences within each theme: models with the same letter are not significantly different whereas models with different letters are significantly different at *α*=0.05. **C-D)** Comparison of the performance of clinical tools to generalist models on HealthBench axes and HealthBench cluster themes, respectively. Statistical significance established via two-way ANOVA and Šidák-corrected post-hoc analysis. In C-D, error bars represent SEM. * $P < 0.05$, ** $P < 0.01$, *** $P < 0.001$, **** $P < 0.0001$.

## Discussion

We present an independent, quantitative comparison of popular clinical AI tools against frontier generalist LLMs. GPT-5 outperformed all models except for Gemini 3 Pro on the MedQA subset of our mini-benchmark, and outperformed all models on the HealthBench subset. Notably, OpenEvidence fell short of its reported 100% accuracy on USMLE-style questions[4]. Our results reveal a consistent pattern in which clinical AI tools lag behind frontier generalist models on numerous dimensions of clinical importance, such as completeness, communication quality, and context awareness. Without knowing the underlying architecture of these systems, it is hard to know why they may underperform generalist models. However research has shown that RAG, which both OpenEvidence and UpToDate Expert AI heavily rely on, can *hurt* model performance if the wrong material is retrieved or if it is poorly integrated by the base model.[7,10,11] Another possibility is simply the strength of frontier LLMs, which excel at the knowledge retrieval and reasoning tasks that characterize many medical questions.

These findings have broader implications for the evolving landscape of clinical AI. As generative models become integrated into routine decision-making, discrepancies between advertised claims and real-world performance introduce avoidable clinical risk. Health systems are increasingly deploying AI-driven assistants in documentation support, guideline lookup, triage, and ambulatory care; settings in which even small reliability deficits can meaningfully affect patient outcomes. Importantly, much of today's AI use is emerging bottom-up, as clinicians and patients experiment with these models on their personal devices well before formal institutional adoption. The consistently superior performance of generalist models suggests that scale, alignment, and cross-domain reasoning may outweigh domain-specific tuning as determinants of medical competency. If this trend continues, the perceived distinction between "clinical LLMs" and "general-purpose LLMs" may become increasingly artificial, with implications for procurement, reimbursement, and regulatory oversight.

This study has several limitations. We were constrained to modest sample sizes due to the need for manual evaluation of OpenEvidence and UpToDate Expert AI. Moreover, standardized benchmarks, while allowing quantitative comparison, have inherent shortcomings.[10] It is possible that recently trained models or tools have been trained on the benchmarks themselves, which were incidentally swept up in their training data. Such concerns underscore the need for open datasets, open weights, and greater transparency surrounding AI systems utilized within clinical medicine. Nonetheless, controlled studies and independent benchmarks of AI systems in medicine are essential to ensure the safe clinical use of these technologies. Perhaps most importantly, however, is the pressing need for actual clinical trials looking at the impact of these technologies with *patient centered outcomes* as their endpoints. Future work should examine performance across broader datasets, contexts, and safety-critical outcomes and the need for regulatory guidance regarding the evaluation of all LLMs in healthcare.

**Acknowledgements:** We would like to acknowledge Nader Mherabi and Dafna Bar-Sagi, Ph.D., for their continued support of medical AI research at NYU. We thank Michael Constantino, Kevin Yie, and the NYU Langone High-Performance Computing (HPC) Team for supporting computing resources fundamental to our work.

**Disclosures:** EKO has equity in Delvi, MarchAI, and Artisight, income from Merck & Co. and Mirati Therapeutics, employment in Eikon Therapeutics, and consulting for Sofinnova Partners and Google. The other authors have no personal, financial, or institutional interest pertinent to this article.

**Author Contributions:** AA and EKO supervised the study. KV, AA, and EKO conceptualized and established the study design. KV designed and performed the LLM evaluations and scoring. MG performed the statistical analysis. KV wrote the initial draft. KV, MG, and AA developed the figures of the manuscript. KV, MG, AA, DAA, YA, and EKO revised and approved the manuscript.

**Funding:** EKO is supported by the National Cancer Institute's Early Surgeon Scientist Program (3P30CA016087-41S1) and the W.M. Keck Foundation. This work was supported by the Institute for Information & Communications Technology Promotion (IITP) grant funded by the Korea government (MSIT) (No. RS-2019-II190075 Artificial Intelligence Graduate School Program (KAIST); No. RS-2024-00509279, Global AI Frontier Lab).

**Supplemental Figures:**

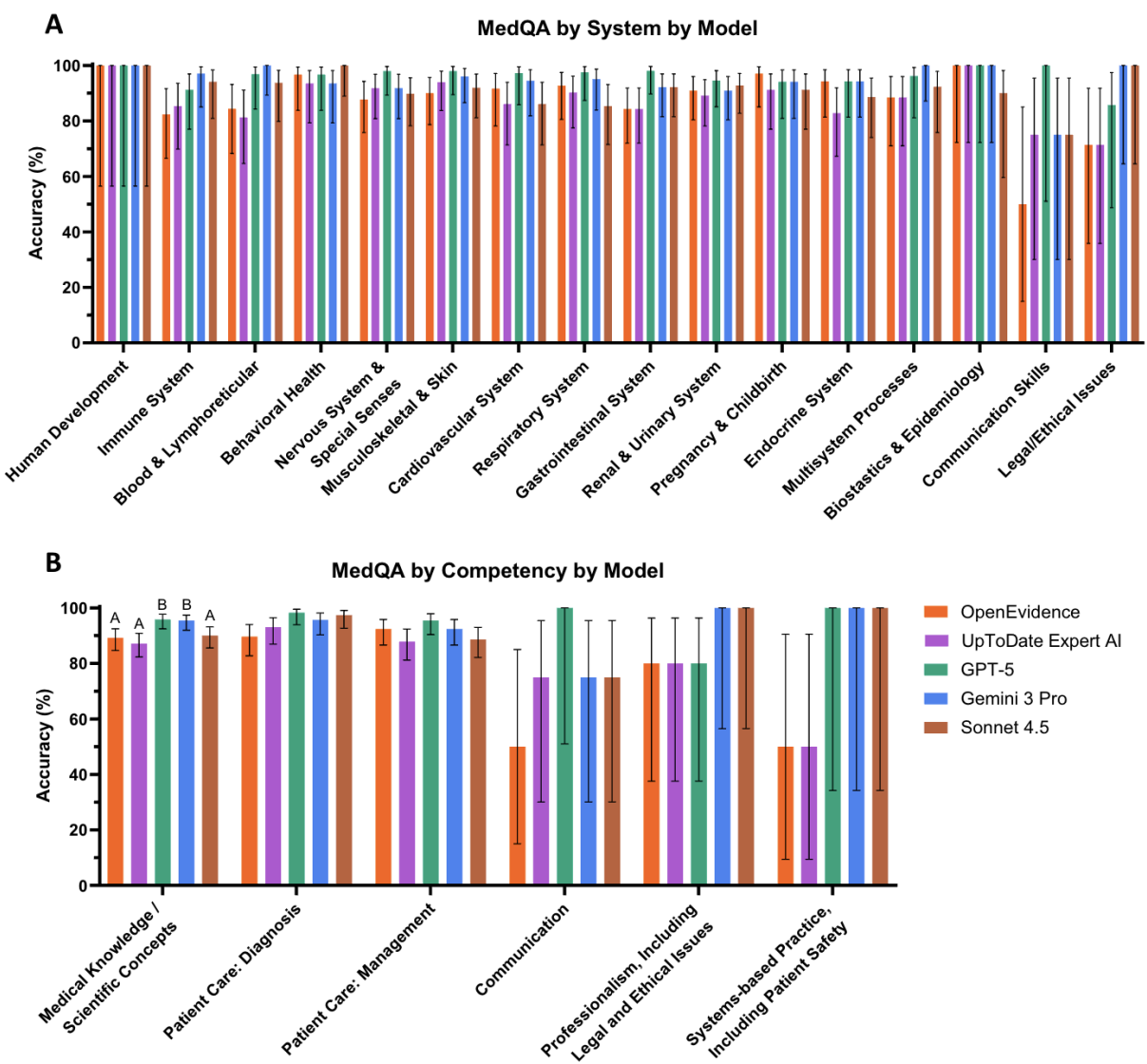

**Supplemental Figure 1.** Performance of each model on USMLE-style questions (MedQA), broken down by **A)** human system category and **B)** USMLE competency category. Statistical significance established via pairwise exact McNemar's tests. Error bars represent 95% confidence intervals, and labeled letters indicate significant differences within each system or within each competency: models with the same letter (or no letter) are not significantly different whereas models with different letters are significantly different at $\alpha$=0.05.

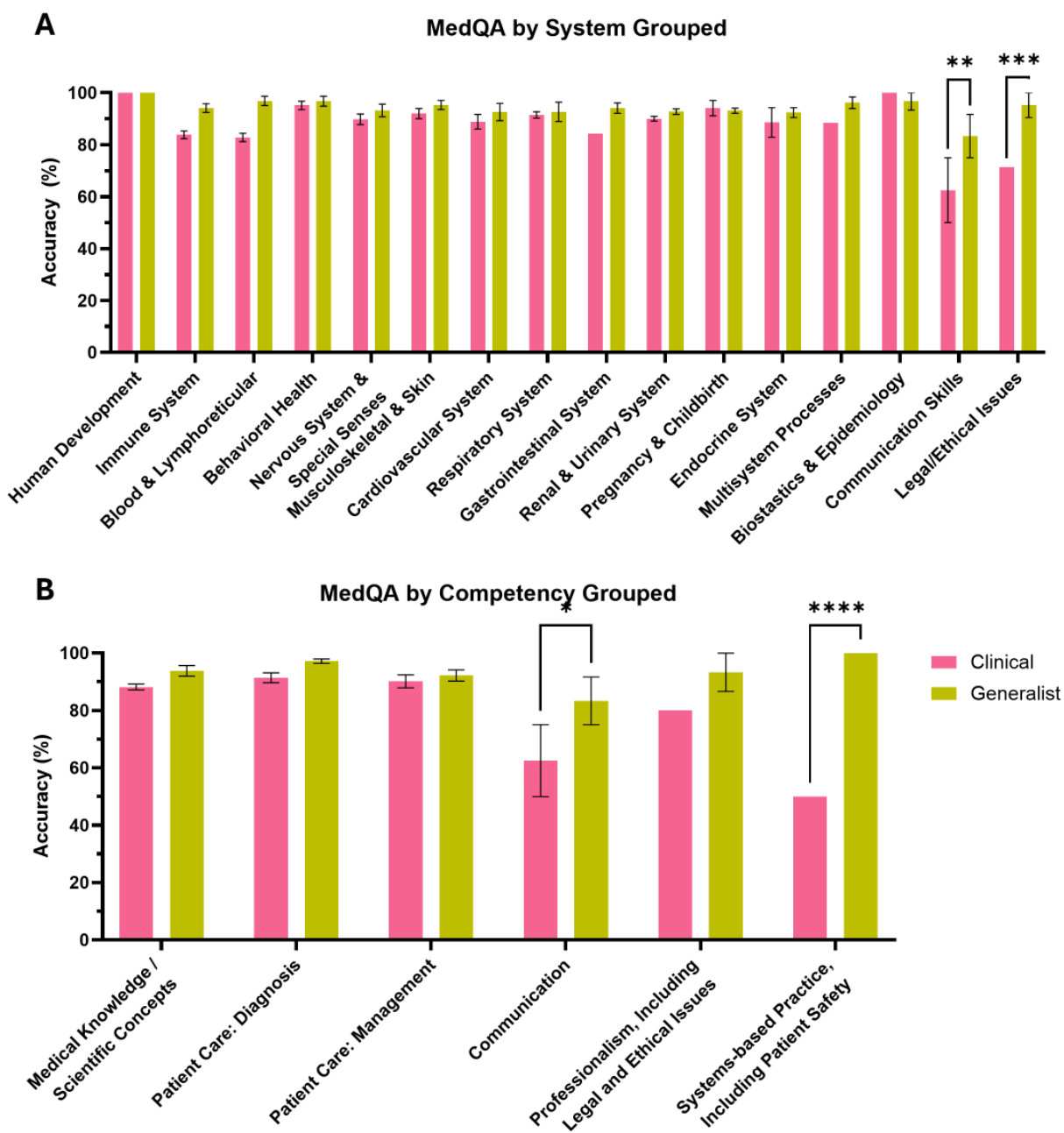

**Supplemental Figure 2.** Performance of clinical versus generalist models on USMLE-style questions (MedQA), broken down by **A)** human system category and **B)** USMLE competency category. Statistical significance established via two-way ANOVA and Šidák-corrected post-hoc analysis. Error bars represent SEM. * $P < 0.05$, ** $P < 0.01$, *** $P < 0.001$, **** $P < 0.0001$.